\documentclass[a4paper,
               ]{jacow}
\usepackage{pdfpages,multirow,ragged2e} %
\makeatletter%
	\ifboolexpr{bool{xetex}}
	 {\renewcommand{\Gin@extensions}{.pdf,%
	                    .png,.jpg,.bmp,.pict,.tif,.psd,.mac,.sga,.tga,.gif,%
	                    .eps,.ps,%
	                    }}{}
\makeatother

\ifboolexpr{bool{xetex} or bool{luatex}} 
 {}                                      
 {\usepackage[utf8]{inputenc}}           

\usepackage[USenglish]{babel}

\ifboolexpr{bool{jacowbiblatex}}%
 {%
  \addbibresource{jacow-test.bib}
  \addbibresource{biblatex-examples.bib}
 }{}
\begin{document}

\title{Scientific QA System with Verifiable Answers}

\author{Adela Ljajić\thanks{adela.ljajic@ivi.ac.rs}\textsuperscript{1},
Miloš Košprdić\textsuperscript{1},
Bojana Bašaragin\textsuperscript{1},
Darija Medvecki\textsuperscript{1},
Lorenzo Cassano\textsuperscript{2}\\
Nikola Milošević\textsuperscript{1}\textsuperscript{2} \\
\textsuperscript{1}Institute for Artificial Intelligence Research and Development of Serbia, Novi Sad, Serbia \\
\textsuperscript{2}Bayer A.G., Berlin, Germany
}

\maketitle

\begin{abstract}
In this paper, we introduce the \textbf{Verif.ai project}, a pioneering open-source scientific question-answering system, designed to provide answers that are not only referenced but also automatically vetted and verifiable. The components of the system are (1) an Information Retrieval system combining semantic and lexical search techniques over scientific papers (PubMed), (2) a Retrieval-Augmented Generation (RAG) module using fine-tuned generative model (Mistral 7B) and retrieved articles to generate claims with references to the articles from which it was derived, and (3) a Verification engine, based on a fine-tuned DeBERTa and XLM-RoBERTa models on Natural Language Inference task using SciFACT dataset. The verification engine cross-checks the generated claim and the article from which the claim was derived, verifying whether there may have been any hallucinations in generating the claim. By leveraging the Information Retrieval and RAG modules, Verif.ai excels in generating factual information from a vast array of scientific sources. At the same time, the Verification engine rigorously double-checks this output, ensuring its accuracy and reliability.
This dual-stage process plays a crucial role in acquiring and confirming factual information, significantly enhancing the information landscape.
Our methodology could significantly enhance scientists' productivity, concurrently fostering trust in applying generative language models within scientific domains, where hallucinations and misinformation are unacceptable.
\end{abstract}

\section{INTRODUCTION}
The introduction of large language models (LLMs) in recent years has marked a transformative phase across numerous sectors, providing advanced capabilities in understanding, generating, and interacting with natural language \cite{openai2023gpt4,jiang2023mistral,bubeck2023sparks,park2023generative,touvron2023llama,katz2023gpt}. Within the scientific realm, these models present an exceptional opportunity to expedite research methodologies, streamline the retrieval of information, and sophisticate the creation of intricate scientific discourse \cite{lewis2020retrieval,ai4science2023impact}. Nevertheless, as these models become more embedded in scientific endeavors, they encounter a pivotal challenge: the phenomena of hallucinations, or the unintended creation of incorrect or misleading content.


In alignment with findings from prior studies on large language models (LLMs) such as those by \cite{Radford2019LanguageMA, Muennighoff2023CrosslingualGT,2022arXiv221105100W}, ChatGPT also encounters the issue of hallucination. The issue of extrinsic hallucination pertains to the creation of unverifiable facts, sourced from its internal memory, across various tasks, without the capability to cross-reference information with external databases. \cite{bang-etal-2023-multitask}.

The issue of hallucinations is particularly critical in scientific settings, where the utmost accuracy and dependability are required. Such occurrences, not only present a barrier to the broader acceptance of LLMs within the scientific community  \cite{boyko2023interdisciplinary}, but also instigate a trust deficit, restricting the full utilization of generative language models due to fears of misinformation. To leverage the full spectrum of advantages offered by these models, it is essential to directly confront and mitigate these concerns, safeguarding the integrity of scientific data.

To address this vital issue, we present \textbf{Verif.ai}, an innovative open-source project designed to minimize the risk of hallucinations in scientific generative question-answering systems. Our strategy employs a multifaceted approach to information retrieval, utilizing both semantic and lexical search methods across extensive scientific databases like PubMed\footnote{\url{https://pubmed.ncbi.nlm.nih.gov/}}. This is complemented by a Retrieval-Augmented Generation (RAG) process, employing the fine-tuned generative model, Mistral 7B, to produce answers with directly traceable references. Additionally, our system employs  an extra layer of fact-checking, or vetting of generated responses.  The verification engine, powered by fine-tuned DeBERTa and RoBERTa models on the SciFACT dataset for the natural language inference task, scrutinizes the congruence of generated claims with their source materials, further solidifying trust in the generated content.

By indicating potential hallucinations and employing advanced hallucination reduction techniques, our system, supported by its open-source framework and the backing of the scientific community, is bridging the trust gap in utilizing LLMs for scientific applications. Through this endeavor, \textbf{Verif.ai} underscores the importance of generating accurate, verifiable information, thereby instilling renewed confidence in the use of LLM-based systems for scientific inquiry.

\section{METHODOLOGY}

\begin{figure*}[!h]
     \centering
     \includegraphics[width=0.7\linewidth]{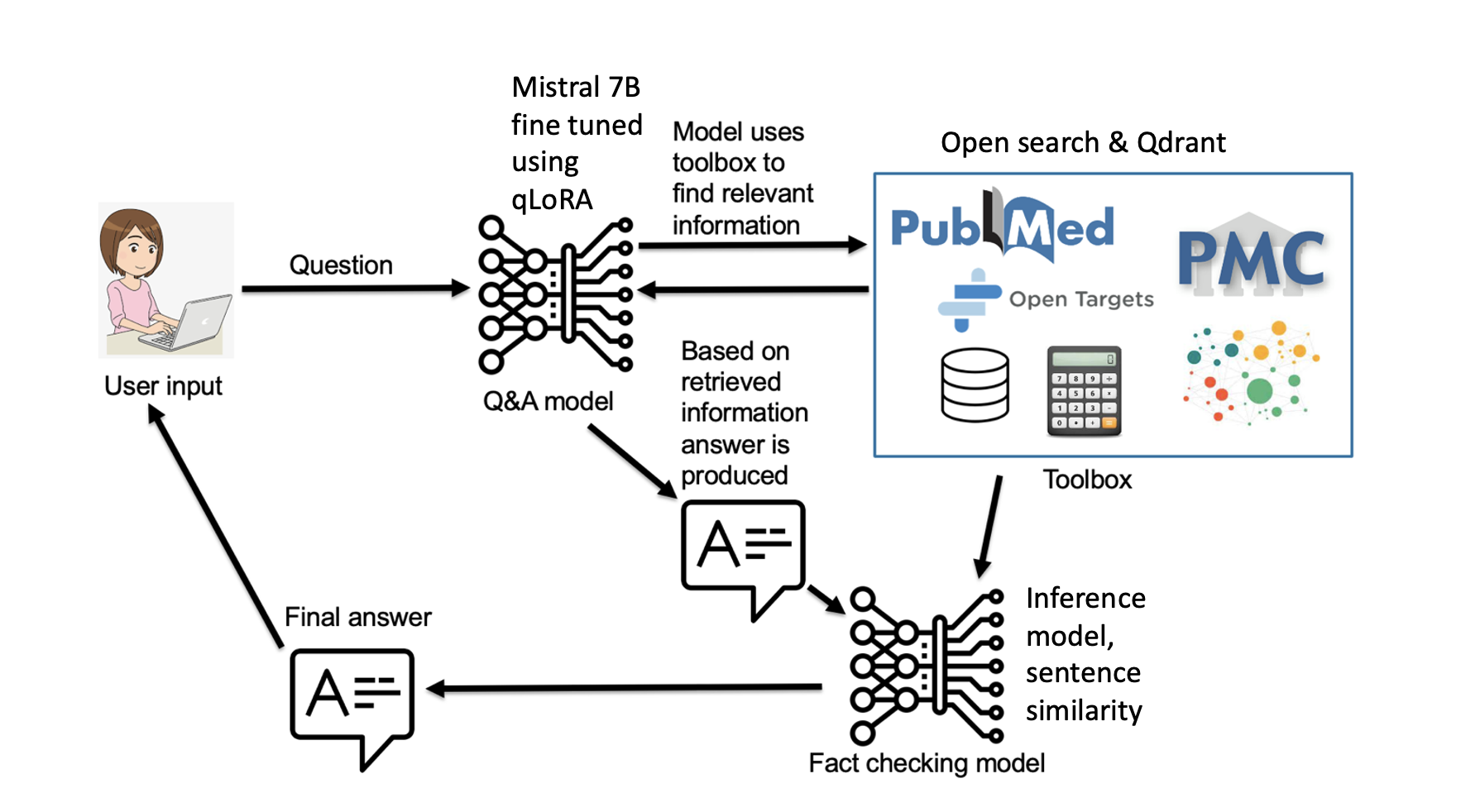}
     \caption{Methodology overview of the \textbf{Verif.ai} project}\label{Fig:InputTransformation}
\end{figure*}

Our methodology employs a toolbox to discover relevant information and provide context to the question-answering system. Currently, the primary component of this toolbox is the information retrieval engine based on indexed documents from PubMed database. The question-answering system utilizes a fine-tuned LLM to generate answers using retrieved documents in context. A fact-checking or verification engine examines the generated answer within the toolbox, identifying any potential hallucinations in the system. The final component of the system is a user interface, enabling users to ask a questions, review answers, and offer feedback functionality, so they can contribute to the improvement of the \textbf{Verif.ai} project. The overview of the methodology is depicted in Figure \ref{Fig:InputTransformation}. In the following subsections, we provide details of the methods envisioned for each of the components.

\subsection{Toolbox and Information Retrieval}

The major component that has been implemented so far in our toolbox is the information retrieval engine. Our information retrieval engine has two components, lexical search and semantic, or vector-based search. The information retrieval component for lexical search is based on OpenSearch\footnote{\url{https://opensearch.org/}}, an open-source engine that was forked from Elasticsearch and is under the Apache 2 license.  The information retrieval component for the semantic search component is based on the Qdrant vector database\footnote{\url{https://qdrant.tech/}}. 

In the compilation of the PubMed corpus from the Medline repository, which encompasses 36,797,469 articles, a deliberate decision was made to exclude articles missing the abstracts. Consequently, the corpus for indexing was refined to 69 \% or exactly to 25,488,790 documents. 
For indexing the content from PubMed articles, we opted to concatenate the title and abstract into a single field, named "text," which serves as the basis for both lexical and semantic searches.
To generate embeddings of the "text" field for vector search, the model trained on the MSMARCO dataset ('sentence-transformers/msmarco-distilbert-base-tas-b') was chosen due to its capability to manage asymmetric searches, such as those involving discrepancies in length between queries and the texts being searched \cite{craswell2021ms}. 

To navigate the tokenization limit of 512 tokens imposed by the selected model for text vectorization, the texts that exceed this threshold underwent a segmentation process. The "text" field is divided into several overlapping segments (with overlap window of 100 tokens) to ensure the preservation of essential information, with each segment being independently indexed to enhance semantic search functionality. This allows for the comprehensive indexing of content that would otherwise be truncated. As a result, the final number of indexed segments is 27,795,286, significantly expanding the scope of searchable content and facilitating more accurate and relevant search results within the indexed corpus.

In the query post-processing, we employ a strategy that integrates the outcomes of both lexical and semantic searches by normalizing their respective retrieval scores to a unified scale range between 0 and 1. This will support direct matches, thereby enhancing the discovery of semantically similar phrases and textual segments where direct text matches are absent. 
Currently, equal weight is accorded to both lexical and semantic searches. However, recognizing the potential for optimization, we plan to adjust these weights by fine-tuning them during the evaluation phase in the future. This adjustment aims to optimize the balance between lexical and semantic search components, enhancing the overall effectiveness and precision of the retrieval system in identifying the most relevant documents. 
The selected documents in the information retrieval phase are then conveyed to the RAG component, responsible for generating the appropriate answer.


\subsection{RAG for Question-Answering with References}

The integration of retrieval components with generative models facilitates the generation of text that is both rich in context and also referenced by the sourced articles. This ensures that the generated claims or responses are not only relevant but also verifiable, drawing directly from the content of the articles retrieved. The RAG framework allows the generative model to reference multiple sources, thereby enriching the response with diverse perspectives and insights. Conversely, it also empowers the model to exclude references to any articles whose content is non-essential or irrelevant to the question, underscoring the model's capacity for critical evaluation and selective synthesis of information.
To integrate the principles of RAG with our methodology, we tested two novel LLMs - Mistral-7B-Instruct-v0.1, a Mistral 7B parameter model with instruction fine-tuning\footnote{\url{https://huggingface.co/filipealmeida/Mistral-7B-Instruct-v0.1-sharded}} and Phi-2 model\footnote{\url{https:huggingface.co/microsoft/phi-2}}. We fine-tuned both models for the task of question-answering with references using a dataset of 10,000 examples containing randomly selected questions from PubMedQA dataset \cite{jin2019pubmedqa}. The answers in the dataset were generated using GPT-3.5 with the most relevant documents from PubMed passed as context. The following prompt was used to generate answers from GPT-3.5:  


\noindent
\qquad \\\fbox{
    \parbox{0.95\linewidth}{
Please carefully read the question and use the provided research papers to support your answers. When making a statement, indicate the corresponding abstract number in square brackets (e.g., [1][2]). Note that some abstracts may appear to be strictly related to the instructions, while others may not be relevant at all.
       }
    }
\\

Fine-tuning was performed using the QLoRA methodology \cite{dettmers2023qlora}. For training we used a rescaled loss, a rank of 64, an alpha of 16, and a LoRA dropout of 0.1, resulting in 27,262,976 trainable parameters. The input to the training has the following structure: the question, retrieved documents (between one and 10 documents), and the answer. 

We then tested the fine-tuned models on the task of answer generation. Using the exactly same input as in the training did not produce the expected results, and therefore, we added the instruction at the beginning of the prompt for both models:


\noindent
\qquad \\\fbox{
    \parbox{0.95\linewidth}{
       prompt = f"""Respond to the Instruction using only the information provided in the relevant abstracts in ```Papers``` below. 

Instruction: \{query\_articles\}

Answer:""" 
    }
}
\\

The beginning instruction was followed by the question asked by the user and 10 relevant documents obtained by querying OpenSearch (lexical search) and Qdrant (semantic search) to retrieve results ranked by this hybrid combination. The instruction is formatted the same way as in the training set. To prompt both models, we use the mentioned template and default parameters with only two differences: we set max\_new\_tokens to 1000 and repetition\_penalty to 1.1.


We made the preliminary generated QLoRA adapter for Mistral available on Hugging Face\footnote{\url{https://huggingface.co/BojanaBas/Mistral-7B-Instruct-v0.1-pqa}}. 

\subsection{Verifying Claims from the Generated Answer}

The aim of the verification engine is to parse sentences and references from the answer generation engine and verify that there are no hallucinations in the answer. Our assumption is that each statement is supported by one or more references. For verification, we compare the XLM-RoBERTa-large model\footnote{\url{https://huggingface.co/xlm-roberta-large}} and DeBERTa model\footnote{\url{microsoft/deberta-v3-large}}, treating it as a natural language inference problem. The selected model has a significantly different architecture than the generation model and is fine-tuned using the SciFact dataset \cite{wadden2020fact}. The dataset is additionally cleaned (e.g., claims were deduplicated, and instances with multiple citations in no-evidence examples were split into multiple samples, one for each reference). The input to the model contains the CLS token (class token), the statement, a separator token, and the joined referenced article title and abstract, followed by another separation token. The output of the model falls into one of three classes: "Supports" (the statement is supported by the content of the article), "Contradicts" (the statement contradicts the article) and "No Evidence" (there is no evidence in the article for the given claim).

The fine-tuned model serves as the primary method for flagging contradictions or unsupported claims. However, additional methods for establishing user trust in the system will be implemented, including presenting to the user the sentences from the abstracts that are most similar to the claim.

\subsection{User Feedback Integration}

The envisioned user interface would present the answer to the user's query, referencing documents containing the answer and flagging sentences that contain potential hallucinations. However, users are asked to critically evaluate answers, and they can provide feedback either by changing a class of the natural language inference model or even by modifying generated answers. These modifications are recorded and used in future model fine-tuning, thereby improving the system.

\section{Preliminary Evaluation}

In this section, we present the results based on our preliminary evaluation. At the time of writing of this article, the project was in the 5rd month of implementation, and we are working on improving our methodology and creating a web application that integrates all the described components.

\subsection{Information Retrieval}

The outcomes derived from employing OpenSearch for lexical search and Qdrant for semantic search were subjected to a qualitative assessment. This examination focused on a subset of indexed articles from the PubMed database, aiming to discern the efficacy and relevance of the search results provided by these distinct methodologies. 
We compared lexical search, semantic search, and a hybrid combination of both lexical and semantic search. We observed that lexical search may perform better when the search terms can be exactly matched in the documents, while semantic search works well with paraphrased text or synonymous terms. Hybrid search managed to find documents containing terms that could be exactly matched, as well as ones that were paraphrased or contained synonyms. While semantic search would also find documents that contained an exact match of the terms, it often happened that they were not prioritized. Hybrid search helped in putting such documents at the top of the search results. Based on several user discussions, we have concluded that users expect the top results to be based on exact matches and later to find relevant documents that do not contain the searched terms.

One of the main challenges in creating hybrid search for large datasets, such as PubMed, is storing the index for the semantic part of the engine. While OpenSearch has support for vector search, by integrating the FAISS vector store, it stores all the vectors in the memory. In case the dataset is large (PubMed contains over 120GB of data with 768 dimensions of embedding vectors), and computational resources are limited, it is necessary to find a performant implementation that would store part of the index on a hard disk. Storing part of the index on a hard drive sacrifices to a certain degree performance, but there are implementations, such as by using memory mapped files \cite{tevanian1987unix} in Qdrant that have acceptable performance. While it requires us to perform two queries and post-process the results ourselves, it enables the implementation of a large and performant index on limited computational resources. 

In our evaluation, we implemented a compression strategy informed by the latest advancements in embedding optimization for vector search, which prioritizes compression over dimensionality reduction for managing large datasets effectively. This approach involved compressing vector embeddings to significantly reduce memory usage 4x by allocating only 1 byte per dimension, thereby retaining 99.99\% of the search quality \cite{CohereCompresion}. Utilizing Qdrant's Scalar Quantization feature allowed us to compress the precision of each dimension from a float 32-bit float to  8-bit unsigned integer. As a result, we not only achieved a substantial reduction in storage and memory requirements but also observed an acceleration in the performance of our semantic search operations, effectively overcoming memory and computational constraints.

\subsection{Answer Generation}

We were able to obtain comparable answers from both models (Mistral 7B and Phi2) when prompting them with a smaller number of documents, but the context length of the models played a deciding role when prompting the models with 10 documents. While Mistral-7B-Instruct-v0.1 can process up to 32,000 tokens of input text, the context length obtained by its instruct fine-tuning, the Phi-2 model can only process up to 2,048 tokens. Since our current goal is to generate answers based on no less than 10 abstracts, this automatically left Phi-2 unfit for our needs. Phi-2 was also prone to over-generation, which was not an issue with Mistral. This issue could only artificially be prevented by lowering the max\_new\_tokens size, but this would also leave the answers unfinished. The combination of these two factors excluded Phi-2 from further testing.

We have manually compared the answers generated by our fine-tuned Mistral-7B-Instruct-v0.1 to answers from GPT-3.5 and GPT-4 on a test set of 50 questions and extracted abstracts. No model showed a clear advantage over the others. The quality, referenced abstracts, and length of the answers varied within each model and among the models. In terms of referenced abstracts, most of the time all three models referenced the same abstracts as relevant. This evaluation indicated that the fine-tuning of the Mistral 7B model improved the model's performance, making the generated answers comparable to those of much larger GPT-3.5 and GPT-4 models for the referenced question-answering task. 

\begin{figure}[!h]
     \centering
     \includegraphics[width=1.0\linewidth]{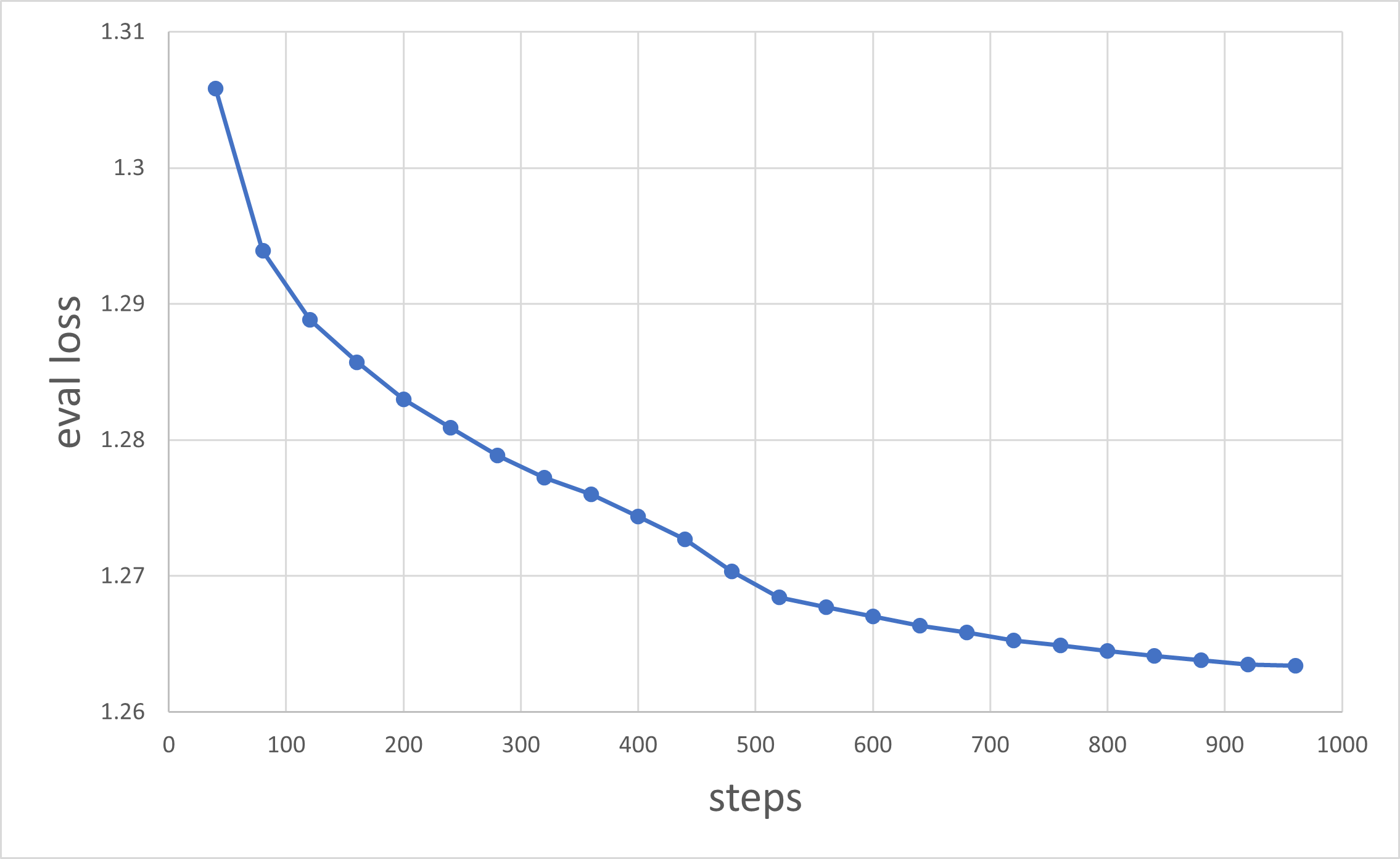}
     \caption{Evaluation loss for fine-tuning of Mistral 7B model on PubMedQA questions with generated and referenced answers}\label{Fig:EvalLoss}
\end{figure}

We have also tested our model on a preliminary output of the information retrieval module which consisted of user queries and 10 relevant documents along with their PubMed IDs. The model showed decreased performance, which seemed to be related to both a different ID format (realistic PMID as opposed to 0-9 numeration of documents in the train set) and a consistently high number of documents. The model was fine-tuned on a varied number of documents for each query, where the highest number of inputs consisted of four documents so this is an expected behavior. Furthermore, in terms of quantitative analysis, fine-tuning of Mistral-7B-Instruct-v0.1 using the original dataset of PubMedQA questions and GPT-3.5 generated answers showed a continuous decrease of evaluation loss, as can be seen in Figure \ref{Fig:EvalLoss}. This behavior leaves space for further improvement so we plan to fine-tune Mistral-7B-Instruct-v0.1 once again using a PMID-like ID format and 10 documents for each query, which could potentially improve the model performance for our purpose. At the time of writing this paper, the dataset is under construction..
 

\subsection{Verification and Hallucination Detection}

The evaluation of the fine-tuned XLM-RoBERTa and DeBERTa model on the SciFact dataset that can be used for hallucination detection can be seen in Table \ref{EntailmentTable}. The model used 10\% of the data for validation and 10\% of the dataset for evaluation (test set). All three sets have homogenous distribution of the classes (36\%:42\%:22\% for NO\_EVIDENCE, SUPPORT and CONTRADICT classes respectively).



\begin{table}[!h]
\scriptsize
\centering
\caption{The evaluation of the entailment model fine-tuned from XLM-RoBERTa-large and DeBERTa-large model using SciFact dataset}
\begin{tabular}{|l|c|c|c|} 
\hline
& \multicolumn{3}{c|}{\textbf{XLM-RoBERTa}}  \\ \hline
& \textbf{Precision}& \textbf{Recall} & \textbf{F1-score}  \\ \hline
\textbf{NO\_EVIDENCE}   & 0.91  & 0.96 & 0.95    \\\hline
\textbf{SUPPORT}   & 0.91  & 0.75 & 0.82   \\ \hline
\textbf{CONTRADICT}   & 0.59  & 0.81 & 0.68    \\ \hline
\textbf{Weighted Avg} & 0.87  & 0.85 & 0.85   \\ \hline
& \multicolumn{3}{c|}{\textbf{DeBERTa}} \\ \hline
\textbf{NO\_EVIDENCE} & 0.88  & 0.86 & 0.87 \\ \hline
\textbf{SUPPORT}  & 0.87  & 0.92 & 0.90  \\ \hline
\textbf{CONTRADICT} & 0.88  & 0.81 & 0.85  \\ \hline
\textbf{Weighted Avg} & 0.88  & 0.88 & 0.88 \\ \hline
\end{tabular}
\label{EntailmentTable}
\end{table}

As can be seen from the table, the models exhibited state-of-the-art performance, surpassing the reported scores in \cite{wadden2020fact} for the label prediction task, and DeBERTa-large model showed superior performance compared to the RoBERTa-large. We use fine-tuned DeBERTa-large model for verification and hallucination detection. We also evaluated the SciFact label prediction task using the GPT-4 model, resulting in a precision of 0.81, recall of 0.80, and an F-1 score of 0.79. Therefore, our models outperformed GPT-4 model in zero-shot regime with carefully designed prompt for label prediction for the claims and abstracts in the SciFact dataset. It is important to note that the SciFact dataset contains challenging claim/abstract pairs, demanding a significant amount of reasoning for accurate labeling. Thus, in a real-use case where answers are generated by Mistral or another generative model, the task becomes easier. We believe that this model provides a good starting point for hallucination detection, as supported by our qualitative analysis of several pairs of generated claims and abstracts, which demonstrated good performance.

However, this model has some limitations. While it is capable of reasoning around negations, detecting contradicting claims, differing in just a few words switching the context of the claim compared to the text of the abstract, proves to be a challenge. Additionally, we observe that neither model handles well situations where numerical values in claims are slightly different from the ones in the abstract.

\section{Conclusion}

In this paper, we present the current progress on the \textbf{Verif.ai} project, an open-source generative search engine with referenced and verifiable answers based on PubMed. We describe our use of OpenSearch and Qdrant to create a hybrid search, an answer generation method based on fine-tuning the Mistral 7B model, and our first hallucination detection and answer verification model based on fine-tuned DeBERTa-large model. However, there are still some challenges to be addressed and work to be done.

LLMs are rapidly developing, and performant, smaller LLMs, with larger context sizes are becoming more available. We aim to follow this development and use the best available open-source model for the task of referenced question-answering. We also aim to release early and collect user feedback. Based on this feedback, we aim to design an active learning method and incorporate user feedback into the iterative training process for both answer generation and answer verification and hallucination detection.

We also aim to improve our answer generation model, by creating better dataset, using GPT4Turbo and manual labelling. We plan to release this dataset in the future.

The model for hallucination detection and answer verification exhibits some limitations when it needs to deal with numerical values or perform complex reasoning and inference on abstracts. We believe that a single model may not be sufficient to verify the abstract well, but it may be the case that a solution based on a mixture of experts may be required \cite{jacobs1991adaptive,jiang2024mixtral}. To build user trust, we aim to offer several answer verification methods, some of which should be based on explainable AI and be easy for users to understand. In the future, this may include, for example, verification based on sentence similarity scores.

Currently, the system is designed for use in the biomedical domain and provides answers based on scientific articles indexed in PubMed. However, we believe that the system can be easily extended to other document formats and become a base for a personal, organizational, or corporate generative search engine with trustworthy answers. In the future, our version may incorporate additional sources, contributing to the trust and safety of the next generation internet.

\section{Availability}

Code created so far in this project is available on GitHub\footnote{\url{https://github.com/nikolamilosevic86/verif.ai}} under AGPLv3 license. Our fine-tuned qLoRA adapter model for referenced question answering based on Mistral 7B \footnote{\url{https://huggingface.co/filipealmeida/Mistral-7B-Instruct-v0.1-sharded}} is available on HuggingFace\footnote{\url{https://huggingface.co/BojanaBas/Mistral-7B-Instruct-v0.1-pqa}}. The verification models are available on HuggingFace\footnote{\url{https://huggingface.co/nikolamilosevic/SCIFACT_xlm_roberta_large}} \footnote{\url{https://huggingface.co/MilosKosRad/DeBERTa-v3-large-SciFact}}. More information on the project can be found on the project website: \url{https://verifai-project.com}.

\section*{Acknowledgment}
The \textbf{Verif.ai} project is a collaborative effort of Bayer A.G. and the Institute for Artificial Intelligence Research and Development of Serbia, funded within the framework of the NGI Search project under Horizon Europe grant agreement No 101069364.

%


\bibliographystyle{vancouver}
\bibliography{cas}
\vspace{12pt}

\end{document}